\documentclass[10pt, conference, letterpaper]{IEEEtran}
\IEEEoverridecommandlockouts
\usepackage{booktabs}
\usepackage{cite}
\usepackage{amsmath,amssymb,amsfonts}
\usepackage{array}
\usepackage{authblk}
\usepackage{algorithmic}
\usepackage{float}
\usepackage{graphicx}
\usepackage{hyperref}
\usepackage{xcolor}
\definecolor{medium-blue}{rgb}{0,0,1}
\hypersetup{colorlinks, urlcolor={medium-blue}}
\usepackage{multirow}
\usepackage{subfiles}
\usepackage{subfigure}
\usepackage{textcomp}
\usepackage{xcolor}
\usepackage{url}

\begin{document}
% \title{Label Leakage and Protection in Vertical Logistic Regression}
\title{Residue-based Label Protection Mechanisms in Vertical Logistic Regression}

\author[$\dagger$]{Juntao Tan}
\author[$\dagger$]{Lan Zhang}
\author[$\ddagger$]{Yang Liu}
\author[$\dagger$]{Anran Li}
\author[$\ddagger$]{Ye Wu}
\affil[$\dagger$]{School of Computer Science and Technology, University of Science and Technology of China, Hefei, China}
\affil[$\ddagger$]{ByteDance Security Research Department, Bejing, China}
\affil[ ]{\{tjt, anranLi\}@mail.ustc.edu.cn, zhanglan@ustc.edu.cn, \{liuyang.fromthu, wuye.2020\}@bytedance.com}

\maketitle

%%%%%%%%%%%%%%%%%%
% Abstract
%%%%%%%%%%%%%%%%%%
\begin{abstract}
% \documentclass[main.tex]{subfiles}
% \begin{document}
Federated learning (FL) enables distributed participants to collaboratively learn a global model without revealing their private data to each other.
Recently, vertical FL, where the participants hold the same set of samples but with different features, has received increased attention. 
% With vertical federated learning being widely used in many applications, its security receives more and more attention. 
This paper first presents one label inference attack method to investigate the potential privacy leakages of the vertical logistic regression model. 
Specifically, we discover that the attacker can utilize the residue variables, which are calculated by solving the system of linear equations constructed by local dataset and the received decrypted gradients, to infer the privately owned labels. 
To deal with this, we then propose three protection mechanisms, \emph{e.g.}, additive noise mechanism, multiplicative noise mechanism, and hybrid mechanism which leverages local differential privacy and homomorphic encryption techniques, to prevent the attack and improve the robustness of the vertical logistic regression model. 
Experimental results show that both the additive noise mechanism and the multiplicative noise mechanism can achieve efficient label protection with only a slight drop in model testing accuracy, furthermore, the hybrid mechanism can achieve label protection without any testing accuracy degradation, which demonstrates the effectiveness and efficiency of our protection techniques.
% both $\mathcal{M}_{add}$ and $\mathcal{M}_{mult}$ can protect label privacy with a minor decrease of model performance, $\mathcal{M}_{hybrid}$ can not only protect label privacy but also can train a performance-lossless global model.

% In this paper, we reveal the vulnerability of the existing training protocol of vertical logistic regression, that is the attacker can steal the private labels by solving a system of linear equations. 
% Then, we propose an additive noise mechanism $\mathcal{M}_{add}$ and a multiplicative noise mechanism $\mathcal{M}_{mult}$, which are both $\epsilon$-LDP. We also present a hybrid mechanism $\mathcal{M}_{hybrid}$ which leverages both random response and homomorphic encryption.  

% \end{document}

\end{abstract}
\begin{IEEEkeywords}
Federated Learning, Homomorphic Encryption, Local Differential Privacy
\end{IEEEkeywords}

%%%%%%%%%%%%%%%%%%
% Introduction
%%%%%%%%%%%%%%%%%%
\section{Introduction}
% \documentclass[main.tex]{subfiles}
% \begin{document}
% In the last decade with the development of big data, cloud computing, and machine learning especially deep learning techniques, the paradigm of extracting knowledge from data as to empower many applications has got been used widely. These applications usually require centralizing the big data in the cloud center so that machine learning models could be trained. But in recent years some laws such as \textit{GDPR}\cite{gdpr} in Europe and \textit{Personal Information Protection Law}\cite{china-law} in China has forbidden large Internet companies to share user data without any restriction. As a result, companies cannot share data with each other and this data are being isolated. So there is a dilemma: on the one hand, companies hope to share data with others to enlarge the number of samples and to enrich feature space, on the other hand, due to  restrictions of lows, the private data should never leave these companies. 

The success of machine learning rests on the availability of massive amount of data. 
However, it limits machine learning's capability to deal with applications where data has been isolated across different organizations and data privacy has been emphasized \cite{gdpr, china-law}, \textit{e.g.}, user's private pictures and videos\cite{zhang2018cloak, zhang2017pic, du2020patronus} captured by mobile phone as well as social relationships\cite{zhang2013message} should not be leaked during model training. 
Federated learning (FL)\cite{mcmahan2017communication, yang2019federated, li2021sample, li2021efficient} is one emerging technology, which enables multiple parties to collaboratively train a machine learning model by iteratively exchanging model parameters between these parties and a centralized server, meanwhile keeping their datasets private. 
There are three types of FL methods according to the distribution of data, which are horizontal federated learning (HFL) \cite{mcmahan2017communication, li2021privacy}, vertical federated learning (VFL) \cite{li2021label, luo2021feature}, and federated transfer learning (FTL) \cite{liu2020secure} respectively.
HFL considers the scenario where each party has data with different sample IDs but shares many common features.
Different from HFL, in VFL, multiple parties handle data with the same sample IDs, but each party has its own feature set.
This is a common phenomenon in financial, e-commerce, and healthcare applications, \emph{e.g.}, two e-commerce companies and a bank which all serve clients from the same city can jointly learn a model by iteratively exchanging intermediate messages between each other to recommend personalized loans for clients based on their online shopping behaviours through VFL. 

% is first proposed in 2016 by Google\cite{mcmahan2017communication}. It is a decentralized training paradigm that can train a global model without each private dataset leaving its device. Furthermore, Yang et al.\cite{yang2019federated} proposes that FL could be categorized into Horizontal Federated Learning (HFL), Vertical Federated Learning (VFL), and Federated Transfer Learning (FTL) respectively. 

% VFL is suited for the scenario that different participants have the same sample space but differ in feature space, so they collaborate with each other to enrich their features and train a global model with performance better than that trained on their own local dataset. Many proposed training protocols on logistic regression \cite{lr-no-third-party}, boosting tree\cite{cheng2021secureboost} and neural network\cite{zhang2020additively} can achieve a lossless model performance compared with the centralized model, by only interchanging intermediate messages with each other. The private dataset is never left. This is why VFL is being widely used in industry because it can improve business for many companies. 

A series of previous efforts have been devoted to designing VFL algorithms, such as logistic regression \cite{lr-no-third-party}, boosting tree\cite{cheng2021secureboost} and neural network\cite{zhang2020additively} via homomorphic encryption\cite{homosurvey, paillier1999public, fontaine2007survey} or multi-party computation techniques\cite{yao1986generate, zhang2013verifiable, jung2014collusion}, for diverse scenarios. 
Despite the wide applications, VFL has an inherent vulnerability that can be leveraged by an adversarial participant to conduct various malicious attacks, \emph{e.g.}, label inference attacks \cite{li2021label, label-infer}, feature inference attacks \cite{luo2021feature} and sample ID attacks \cite{liu2020asymmetrical}.
% A thorough study of this vulnerability is imperative for the wide applications of VFL
% its security is also been investigated by many researchers\cite{li2021label, luo2021feature, weng2020privacy, liu2020asymmetrical, label-infer}. 
Specifically, for label inference attacks, Li et al.\cite{li2021label} proposes a label-uncovering method which uses the norm of the communicated gradients between the parties as well as a protection technique that perturbs the gradients randomly before communication.
Besides, Chong et al. \cite{label-infer} discover that the bottom model structure and the gradient update mechanism of VFL can be exploited by a malicious participant to gain the power to infer the privately owned labels.
For feature inference attacks, Luo et al.\cite{luo2021feature} presents several feature inference attack methods in the prediction stage of several VFL models, \emph{e.g.}, the linear model, the tree model, and the neural network model.
% The authors of \cite{weng2020privacy} study the security of many widely used federated learning frameworks and show their vulnerability in the case of collusion between attacker and third party. 
Besides, the work \cite{weng2020privacy} considers that an honest-but-curious adversary can infer private training data from the legitimately received information in the case of collusion between attacker and third party. 
Furthermore, for sample ID attacks, Yang et al. \cite{liu2020asymmetrical} proposes the notion of asymmetrical VFL and leverages the standard private set intersection protocol to achieve the asymmetrical ID alignment phase in an asymmetrical VFL system to protect sample IDs.
% Furthermore, the privacy of sample IDs is also been studied by \cite{liu2020asymmetrical}, which shows that a participant with more IDs can steal IDs of another. 

Though these methods can reveal a variety of VFL vulnerabilities, their assumptions are impractical in real VFL applications.
For instance, the work \cite{li2021label} assumes that the categorical distribution of the training samples is unbalanced. 
Meanwhile, the work  \cite{luo2021feature} assumes the attacker controls the whole trained VFL model parameters, which is contrary to the settings of the VFL protocol, and the attacking method in \cite{weng2020privacy} can only work when one participant colludes with third party. 
To this end, in this paper, we focus on the vulnerability discovery and privacy protection of currently widely used vertical logistic regression protocol \cite{lr-no-third-party} without any impractical assumptions. 
Specifically, we discover that an attacker can construct a system of linear equations by its local dataset and the received decrypted gradients, to solve the residue variables and further to infer the private labels owned by other participant. This is a serious data privacy breach for the vertical federated learning system, so we propose three residue protection mechanisms, \emph{e.g.}, the additive noise mechanism, the multiplicative noise mechanism, and a hybrid mechanism that leverages local differential privacy and homomorphic encryption techniques simultaneously, to prevent such an attack. As a result, we can improve the robustness of the vertical logistic regression model.

%Specifically, we discover that the attacker can utilize the residue variables, which are calculated by solving a system of linear equations constructed by local dataset and the received decrypted gradients, to infer the privately owned labels. Moreover, we propose three protection mechanisms, \emph{e.g.}, the additive noise mechanism, the multiplicative noise mechanism, and the hybrid mechanism which leverages local differential privacy (LDP) and homomorphic encryption techniques, to prevent the attack and improve the robustness of the vertical logistic regression model.%

% The attacker can utilize the solved residue variable to infer true labels. And furthermore, we also propose three different defending mechanisms to improve the robustness of the training protocol.

% So in this paper, without any impractical assumptions, we reveal that the training protocol of vertical logistic regression first proposed by \cite{lr-no-third-party} is not secure. 
The main contributions of this work are summarized as follows:
\begin{itemize}
\item We identify an effective label inference method that uses the calculated residue variables from the constructed linear system based on local datasets and the decrypted gradients to infer private labels, without actually decrypting the residues.
\item We propose two computation efficient residue protection mechanisms as well as a hybrid residue protection mechanism. 
These two computation efficient mechanisms utilize additive noise and multiplicative noise to mask residues respectively to achieve label protection, and they satisfy the property of $\epsilon$-LDP in each training round. In addition, the hybrid residue protection mechanism uses the local differential privacy and homomorphic encryption methods to increase the batch size to protect private labels without any testing accuracy degradation.
    % \item We discover that when the number of attacker's features is larger than the mini-batch size, the attacker can construct a system of linear equations by its local dataset and the decrypted gradients, and further solve the residue variables without actually decrypting it to infer private labels.
    % \item We propose two different computation efficient label protection mechanisms which use additive noise and multiplicative noise to mask residues separately, and prove that they both satisfy  $\epsilon$-LDP. 
    % \item We also propose a hybrid label protection mechanism which leverages both local differential privacy and homomorphic encryption to enlarge mini-batch size in each training iteration to prevent label leakage, and show that it can achieve a lossless model performance. 
    \item We conduct extensive experiments on four public datasets to evaluate the three  proposed residue protection mechanisms, and the results demonstrate that our protection techniques are both effective and efficient.

    %\item We discover the vulnerability of training protocol in \cite{lr-no-third-party}, which is that when the number of attacker's features is larger than mini-batch size, the private label will be leaked.
    %\item We propose two $\epsilon$-LDP mechanisms to protect the private label, which is computation efficient compared with \cite{lr-no-third-party}.
    %\item We also propose a hybrid mechanism and show that it can not only protect label privacy but also can train a global model with lossless performance.
    %\item We conduct extensive experiments on four datasets to evaluate the three mechanisms, and the results demonstrate the effectiveness of these mechanisms. 
\end{itemize}

% \end{document}

\label{sec:intro}
% \subfile{intro}

%%%%%%%%%%%%%%%%%%
% preliminaries
%%%%%%%%%%%%%%%%%%
\section{Background and Preliminaries}
% \documentclass[main.tex]{subfiles}
% \begin{document}

\subsection{Logistic Regression}
% Consider the task of learning a mapping $f(\boldsymbol{W}) = \mathcal{X} \rightarrow \mathcal{Y}$ with parameters $\boldsymbol{W}$, that maps an input space $\mathcal{X}$ to an output space $\mathcal{Y}$. 
The supervised machine learning aims to learn a mapping $f(\boldsymbol{W}) = \mathcal{X} \rightarrow \mathcal{Y}$ from an input space $\mathcal{X}$ to an output space $\mathcal{Y}$, where $\boldsymbol{W}$ is the model parameters. 
To be concrete, for a training dataset $T=\{(x_1, y_1), (x_2, y_2), \cdots, (x_N, y_N)\}$, the supervised machine learning tries to minimize the following loss function:

\begin{equation*}
    \mathcal{L} = \frac{1}{N} \sum_{i=1}^N l(f(x_i;\boldsymbol{W}), y_i) + \lambda \Omega(\boldsymbol{W}),
\end{equation*}

\noindent where $l(\cdot)$ is the loss function and $\Omega(\cdot)$ is the regularization term which is used to reduce model complexity in order to prevent overfitting, and $\lambda$ represents regularization parameter, which controls the trade-off between the empirical loss and the regularization loss.

Logistic regression (LR) is a binary supervised machine learning model with output space $\mathcal{Y} \in \{0, 1\}$. In LR, the non-linear mapping function is 

\begin{equation*}
    f(\boldsymbol{W}) = \sigma(\boldsymbol{W}^T \boldsymbol{x}), \textrm{where} \; \sigma(z) = \frac{1}{1 + e^{-z}},
\end{equation*}

\noindent which is the $\mathtt{sigmoid}$ function. And the loss function is 

\begin{equation} \label{eq:loss}
    \mathcal{L} = -\frac{1}{N} \sum_{i=1}^N y_i \log(f(x_i)) + (1-y_i) \log(1-f(x_i))  + \lambda \Omega(\boldsymbol{W}),
\end{equation}

\noindent where $\Omega(\boldsymbol{W})$ can take the form of $L_1$ norm or $L_2$ norm of $\boldsymbol{W}$. And gradients are computed as

\begin{equation} \label{eq:grad}
    \frac{\partial \mathcal{L}}{\partial \boldsymbol{W}} = -\frac{1}{N} \sum_{i=1}^N (y_i - f(x_i)) x_i .
\end{equation}

\noindent For a trained model and a testing data sample $x_{ pred}$, we can get the probability of this sample being classified as positive as $f(x_{pred};\boldsymbol{W} )$.

\subsection{Homomorphic Encryption}

Different from the conventional symmetric or asymmetric encryption, homomorphic encryption (HE) is a special kind of cryptosystem which can support arithmetic computation on ciphertext\cite{homosurvey, paillier1999public, fontaine2007survey}. It can ensure that after decryption the computation result on the ciphertext is the same as the computation result on the plaintext.

Additive homomorphic encryption belongs to partially homomorphic cryptosystem, which can support computation on an logical circuit with infinite depth of addition gates, mainly has the following two properties, 
\begin{equation*}
    \begin{aligned}
        \langle m_1 \rangle \oplus \langle m_2 \rangle  = \langle m_1 + m_2 \rangle \\
        m_1 \star \langle m_2 \rangle = \langle m_1 \times m_2 \rangle
    \end{aligned}
\end{equation*}

\noindent where $\langle m_1 \rangle$ and $\langle m_2 \rangle$ represents two ciphertexts, $\oplus$ stands for addition operation on ciphertext, and $\star$ represents multiplication operation on ciphertext. When the result of the addition or multiplication operation on the ciphertext is decrypted, it is the same as the result of the operation on the plaintext.

\subsection{Local Differential Privacy}
Differential privacy (DP) is a privacy-preserving technique that is used to obscure the output of an oracle.
Local differential privacy (LDP) \cite{ldpcollecting2016, ldpsurvey2020, ldpwang2019}, compared with DP, the main difference is that the role to add noise is the owner of private data, not the central aggregator, which can further reduce the risk of privacy leakage.

We say that a randomized algorithm $\mathcal{M}$ satisfies $\epsilon$-LDP if and only if for any input $t$ and $t'$ in the input space, and for $t^{\ast}$
in the output space, we have:

\begin{equation*}
    \Pr[\mathcal{M}(t) = t^{\ast}] \leq e^{\epsilon} \cdot \Pr[\mathcal{M}(t') = t^{\ast}],
\end{equation*}

\noindent where $\Pr[\cdot]$ denotes probability and $\epsilon$ stands for privacy budget in the context of differential privacy. With a lower privacy budget, we can provide a stronger privacy guarantee.

% \end{document}

%%%%%%%%%%%%%%%%%%
% Problem statement
%%%%%%%%%%%%%%%%%%
\section{Problem Statement}
% \documentclass[main.tex]{subfiles}
% \begin{document}
In this section, we first introduce the system model of vertical federated learning in Sec.\ref{system-model}. Then, in Sec.\ref{adversary-model}, we discuss the adversary model. Finally, we give a formal definition of the research problem in Sec.\ref{problem-formulation}.

\subsection{System Model} \label{system-model}
In a vertical federated learning system, without loss of generality, we suppose there are two participants. One participant's private dataset contains both features and labels, called Bob, being the \textit{active party}, the other's dataset contains only features, called Alice, being the \textit{passive party}.

The private dataset of Alice can be represented as $\mathcal{D}_A = \{x^A_1, x^A_2 \cdots, x^A_N\}$, where $x^A_i \in \mathcal{X}^{d^A}$ and $d^A$ is the number of Alice's features. Bob's private dataset is denoted as $\mathcal{D}_B = \{(x^B_1, y_1), (x^B_2, y_2), \cdots, (x^B_N, y_N)\}$, where $x^B_i\in \mathcal{X}^{d^B}$ and $d^B$ is the number of Bob's features, and $y_i \in \mathcal{Y}$, in our case of vertical logistic regression, $\mathcal{Y} = \{0, 1\}$. 

Alice and Bob collaboratively train a vertical logistic regression model without the assistance of a trusted third party. We assume that the \textit{private set intersection} (PSI) procedure before model training, which can be implemented via protocols in \cite{psi01, psi02, psi03},  has been completed. After model training, Alice and Bob will obtain the model parameters associated with its feature space, which 
are $\boldsymbol{W}^A \in \mathbb{R}^{d^A}$ and $\boldsymbol{W}^B \in \mathbb{R}^{d^B}$ respectively. For a given testing sample $(x^A_{pred}, x^B_{pred})$, Alice and Bob can get the prediction result $\hat{y} = f(x^A_{pred}, x^B_{pred}; \boldsymbol{W}^A, \boldsymbol{W}^B)$ via a collaborative inference protocol\cite{yang2019federated, lr-no-third-party}.

\subsection{Adversary Model} \label{adversary-model}
In the two-party vertical federated learning system, we assume that one party, the passive party Alice, is an honest-but-curious party. Alice strictly follows the collaborative training protocol, but she tries her best to infer some valuable information through intermediate messages. For example, Alice may want to infer Bob's model parameters, features, and labels, etc. We do not assume Alice to be a malicious party because the goal of vertical federated learning is to get a better global model, if Alice tries to attack the collaborative protocol itself, then she will not get a model with good performance, and that is inconsistent with Alice's goal.

The adversary model can be characterized by two aspects, which are adversary's goal, adversary's knowledge respectively.

\begin{itemize}
    \item \textbf{Adversary's goal}. In the vertical logistic regression setting, Alice's goal is to infer the private labels of Bob's dataset via the transferred intermediate messages.
    
    \item \textbf{Adversary's knowledge}. Alice's knowledge about the system include 1) the private dataset $\mathcal{D}_A$; 2) feature space $\mathcal{X}^{d_A}$; 3) number of her features $d^A$; 4) partial model parameters $\boldsymbol{W}^A$ associated with $\mathcal{X}^{d_A}$.
        % \begin{itemize}
        %     \item[-] the private dataset $\mathcal{D}_A$;
        %     \item[-] feature space $\mathcal{X}^{d_A}$;
        %     \item[-] number of her features $d^A$;
        %     \item[-] partial model parameters $\boldsymbol{W}^A$ associated with $\mathcal{X}^{d_A}$.
        % \end{itemize}
    
\end{itemize}

\subsection{Problem Formulation} \label{problem-formulation}
For the widely used training protocol of vertical logistic regression, at each iteration, Bob chooses a mini-batch $\mathcal{B}$ and calculates the corresponding linear predictions, losses and gradients. 
% For each sample $x_i^A$, $x_i^B$, 
Bob can compute residue as $r_i=y_i - f(x_i)$ and sends the encrypted form, which is $\langle r_i \rangle$, to Alice. According to Eq.\ref{eq:grad} Alice can calculate encrypted gradients on the encrypted residue and update her model parameters via the gradient decrypted by Bob. For a more detailed description of the training protocol, please refer to \cite{lr-no-third-party}.

After receiving the encrypted residue vector $\langle \boldsymbol{r} \rangle$ and decrypted gradient $\boldsymbol{g}^A$ on the current mini-batch $\mathcal{B}$ from Bob, Alice can construct the following equations:

\begin{equation}
    \label{eq:linear-equations}
    \begin{cases}
    X^A_{1, 1} \langle r_1 \rangle + X^A_{2, 1} \langle r_2 \rangle + \cdots + X^A_{\mathcal{B}, 1} \langle r_{\mathcal{B}} \rangle&= g_1\\ 
    X^A_{1, 2} \langle r_1 \rangle + X^A_{2, 2} \langle r_2 \rangle + \cdots + X^A_{\mathcal{B}, 2} \langle r_{\mathcal{B}}\rangle &= g_2 \\
    &\vdots\\
    X^A_{1, d^A} \langle r_1 \rangle + X^A_{2, d^A} \langle r_2 \rangle + \cdots + X^A_{\mathcal{B}, d^A} \langle r_{\mathcal{B}}\rangle &= g_{d^A}
    \end{cases}
\end{equation}

\noindent The linear equations in Eq.\ref{eq:linear-equations} with $|\mathcal{B}|$ unknowns and $d^A$ equations can be represented as a vectorized form as $(X^A_{\mathcal{B}})^{\mathsf{T}}  \langle \boldsymbol{r} \rangle = \boldsymbol{g}^A$.  Since in many practical applications, many companies and institutes have thousands of features and they prefer a small mini-batch size in order to get a faster convergence rate, so the number of Alice's features is often larger than the mini-batch size, as a result, we can derive that 

\begin{equation}
    \label{eq:rank}
    rank \left((X^A_{\mathcal{B}})^{\mathsf{T}} \right) = rank \left((X^A_{\mathcal{B}})^{\mathsf{T}}, \boldsymbol{g}^A \right) = |\mathcal{B}|.
\end{equation}

\noindent Then the linear system in Eq.\ref{eq:linear-equations} has one and only one solution so that Alice can get the true values of residues by solving the linear system, without actually decrypting the ciphertext. And as mentioned before, in logistic regression model, the ground truth label $y_i$ lies in the space of $\mathcal{Y} = \{0, 1\}$, the residue $r_i$ is the subtraction of the ground truth label $y_i$ and predicted value of $x_i$, since we use $\mathtt{sigmoid}$ as activation function,  we have $0< f(x_i) < 1$. Then if Alice knows that $r_i > 0$, she can infer the ground truth label $y_i$ equals 1, and if $r_i < 0$, she can get  that the true label is 0. In such a scenario, Alice could steal Bob's private label without even interrupting the training protocol, and this is a serious privacy breach for Bob. 

So in this paper, we intend to leverage privacy-preserving techniques, especially local differential privacy and additive homomorphic encryption, to design residue protecting mechanisms to prevent Bob's private label from being breached.

% \end{document}

%%%%%%%%%%%%%%%%%%
% Main methods
%%%%%%%%%%%%%%%%%%
\section{Residue Protecting Mechanisms}
% \documentclass[main.tex]{subfiles}
% \begin{document}
In Sec.\ref{problem-formulation}, we talk about that, by solving the linear equations, Alice can get the residues, and further she can infer Bob's private labels. So in this section, we try to design residue protecting mechanisms to prevent label leakage. Specifically, in Sec.\ref{mechanism-add} we introduce the additive noise mechanism and prove that it is $\epsilon$-LDP. Then we provide another $\epsilon$-LDP mechanism via multiplicative noise in Sec.\ref{mechanism-mult}. Finally, we devise a hybrid mechanism in Sec.\ref{mechanism-hybrid}, which leverages random response and additive homomorphic encryption, and it is proved performance-lossless.

\subsection{Additive Noise Mechanism $\mathcal{M}_{add}$} \label{mechanism-add}

In an additive noise mechanism, Bob tries to add well-crafted noise to the original residue set, and sends the masked residue to Alice so to make it hard for her to infer the ground-truth value.

For any $r_i$, $r_j$ in residue set $R$, we have $-1<r_i<1$ and $-1<r_j<1$, so the $L_1$ sensitivity of function $f(r) = r$ on set $R$ is $\Delta = \max_{r_i, r_j \in R} \left \| r_i - r_j  \right \|_1= 2$. Then the additive noise mechanism can be written as

\begin{equation} \label{eq:add}
    \mathcal{M}_{add}(r) = r + \text{Lap} \left( \frac{2}{\epsilon} \right),
\end{equation}

\noindent where $\text{Lap}(\cdot)$ represents Laplace distribution. This mechanism satisfies $\epsilon$-LDP and we can prove it as

\begin{equation}
    \begin{aligned}
    \frac{\Pr[\mathcal{M}_{add}(r_i) = z]}{\Pr[\mathcal{M}_{add}(r_j) = z]} &= \frac{\exp \left(- \frac{\epsilon |z - r_i|}{\Delta}\right)}{\exp\left(-\frac{\epsilon |z-r_j|}{\Delta}\right)} \\
    &= \exp\left(\frac{\epsilon(|z - r_i| - |z-r_j|)}{\Delta}\right) \\
    &\leq \exp\left( \frac{\epsilon |r_i - r_j|}{\Delta}\right) \\
    & \leq \exp(\epsilon).
    \end{aligned}
\end{equation}

So by leveraging additive mechanism $\mathcal{M}_{add}$, Bob can send the masked residue  to Alice without any encryption, this can help improve the training efficiency compared with the original two-party training protocol in \cite{lr-no-third-party}. And the privacy of residue is guaranteed by the privacy budget $\epsilon$. The impact of the additive noise on the global model accuracy and AUC score is evaluated in Sec.\ref{exp:add}.

\begin{table*}[ht!]
    \centering
    \caption{Training protocol of hybrid mechanism $\mathcal{M}_{hybrid}$ based on \\random response and additive homomorphic encryption}
     \label{tab:hybrid}
    \begin{tabular}{m{1cm}|p{6cm}|p{4cm}|m{2cm}}
        \toprule
        \textbf{Steps} & \textbf{Active party Bob} & \textbf{Passive party Alice} & \textbf{Transmitted messages from Bob to Alice} \\
        \midrule
        Step 0  & Generate paillier\cite{paillier1999public} key pairs and send public key to Alice. & Receive public key.  & public key \\
        \hline
        Step 1 & Choose subset $S$ with indices $\mathcal{B}$ from dataset $X^B$ and generate a binary indicator vector $\boldsymbol{m}$ with the same length as $S$. Set $q|S|$ items of $\boldsymbol{m}$ to 1 and the rest to 0 randomly, and send the output of random response which is $RR(\boldsymbol{m})$, and indices $\mathcal{B}$ to Alice. & Calculate partial linear prediction as $l^A = X^A_{\mathcal{B}} \boldsymbol{W}^A RR(\boldsymbol{m})$ and send it back to Bob. & $RR(\boldsymbol{m})$, $\mathcal{B}$\\
        \hline
        Step 2 & Calculate partial linear prediction as $l^B = X^B_{\mathcal{B}} \boldsymbol{W}^B RR(\boldsymbol{m})$ locally and receive $l^A$ from Alice, combine $l^A$ and $l^B$ and then compute loss and residue $\boldsymbol{r}$ according to Eq.\ref{eq:loss}.  Select $0< k \leq |S|$ non-zeros items from  $\boldsymbol{r}$ to get $\boldsymbol{r}_1$, and set all other items to 0 to get $\boldsymbol{r}_2$, then encrypt $\boldsymbol{r}_1 \cup \boldsymbol{r}_2$ and send it to Alice. & Compute encrypted gradient as $ \langle \boldsymbol{g}^A \rangle = -\frac{1}{|\mathcal{B}|}(X^A_{\mathcal{B}})^{\mathsf{T}} \langle \boldsymbol{r}_1 \cup \boldsymbol{r}_2 \rangle $,  then use noise $\xi$ to mask the gradient and send $ \langle \boldsymbol{g}^A + \xi \rangle$  back to Bob. & $\langle \boldsymbol{r}_1 \cup \boldsymbol{r}_2 \rangle$\\
        \hline
        Step 3 &First, decrypt the encrypted masked gradient and sent $\boldsymbol{g}^A + \xi$ to Alice and second, compute gradient as $\boldsymbol{g}^B = -\frac{1}{|\mathcal{B}|} (X^B_{\mathcal{B}})^{\mathsf{T}} (\boldsymbol{r}_1 \cup \boldsymbol{r}_2)$ locally and use $\boldsymbol{g}^B$ to update parameter $\boldsymbol{W}^B$.& Remove the mask $\xi$ to get the true gradient $\boldsymbol{g}^A$ and then use it to update parameter $\boldsymbol{W}^A$. & $\boldsymbol{g}^A + \xi$\\
        \bottomrule
    \end{tabular}
\end{table*}

\subsection{Multiplicative Noise Mechanism $\mathcal{M}_{mult}$} \label{mechanism-mult}

In the multiplicative noise mechanism, Bob generates noise from a Laplace distribution and multiplies it with the residue to get  $\mathcal{M}_{mult}$. By defining two clipping methods $\mathtt{clip}_1$ and $\mathtt{clip}_2$, we can prove that $\mathcal{M}_{mult}$ is $\epsilon$-LDP. We give a more detailed explanation below.

Firstly, we define the first clipping method as

\begin{equation} \label{eq:clip1}
    \mathtt{clip}_1(r) = 
    \begin{cases}
    b_1, &|r| \leq b_1, \\
    r, & \text{otherwise}
    \end{cases}
\end{equation}

\noindent where $b_1$ is the clipping bound of $\mathtt{clip}_1$. In this way, we can ensure that the $L_1$ sensitivity of function $f(r) = \frac{1}{r}$ on the residue set $R$ is $\Delta = \max_{r_i, r_j \in R} \left \| \frac{1}{r_i} - \frac{1}{r_j} \right \|_1 = \frac{2}{b_1}$. 

Then, the second clipping method is defined as

\begin{equation} \label{eq:clip2}
    \mathtt{clip}_2(\mathcal{M}_{mult}(r)) = 
    \begin{cases}
    b_2, & |\mathcal{M}_{mult}(r)|\geq b_2, \\
    \mathcal{M}_{mult}(r), & \text{otherwise}
    \end{cases}
\end{equation}

\noindent where $b_2$ is the clipping bound of $\mathtt{clip}_2$. So we can derive our multiplicative mechanism as the following

\begin{equation} \label{eq:mult}
    \mathcal{M}_{mult}(r) = r \cdot \text{Lap}\left(\frac{2b_2}{b_1 \epsilon}\right).
\end{equation}

\noindent Still, we can prove that Eq.\ref{eq:mult} satisfies $\epsilon$-LDP as

\begin{equation}
    \begin{aligned}
    \frac{\Pr[\mathcal{M}_{mult}(r_i)=z]}{\Pr[\mathcal{M}_{mult}(r_j)=z]} &= \frac{\exp\left(-\frac{\epsilon|\frac{z}{r_i}|}{b_2 \Delta}\right)}{\exp\left(-\frac{\epsilon|\frac{z}{r_j}|}{b_2 \Delta} \right)} \\
    &= \exp\left( \frac{\epsilon(|\frac{z}{r_j}| -|\frac{z}{r_i}|)}{b_2 \Delta }\right) \\
    & \leq \exp\left( \frac{\epsilon |z|}{b_2}\right) \\
    & \leq \exp(\epsilon).
    \end{aligned}
\end{equation}

So Bob can use this mechanism to protect the residues and avoid the time-consuming encryption operation at each iteration. Still, the privacy of residues is guaranteed by privacy budget $\epsilon$. In Sec.\ref{exp:mult}, we evaluate the impact of parameters $b_1$ and $b_2$ on the global model performance, we also provide the comparison between $\mathcal{M}_{add}$ and $\mathcal{M}_{mult}$.

\subsection{Hybrid Mechanism $\mathcal{M}_{hybrid}$ based on LDP and HE} \label{mechanism-hybrid}

In previous Sec.\ref{mechanism-add} and Sec.\ref{mechanism-mult}, we have talked about how Bob can obtain an $\epsilon$-LDP mechanism via additive and multiplicative noise. The $\mathcal{M}_{add}$ and $\mathcal{M}_{mult}$ mechanisms are computational efficient because there are no encryption and decryption operations in it, but the extra noise can decrease the model performance. So in this section, we propose a hybrid mechanism $\mathcal{M}_{hybrid}$, which leverages random response and additive homomorphic encryption to implement a lossless training protocol, which means that the global model accuracy and AUC score are the same as the one obtained from the conventional centralized model training.

\subsubsection{Training Protocol of $\mathcal{M}_{hybrid}$}

Table \ref{tab:hybrid} shows the details of the training protocol, which is based on random response\cite{warner1965randomized} and additive homomorphic encryption. The protocol shows  just one iteration of a training algorithm, usually, in practical application, such a protocol is often iterated many times to get a model with good performance.

Concretely, in Step 1, we use random response to obscure items in indicator vector $\boldsymbol{m}$ with  probability $p = \frac{e^{\epsilon}}{1 + e^{\epsilon}}$ to be unchanged and with probability $1-p$ to be flipped. To this end, we can prove that the random response mechanism is $\epsilon$-LDP \cite{dwork2014algorithmic}. Furthermore , in order to prevent Bob's private label from being breached, the parameters in Step 1 need to satisfy the requirements in Eq.\ref{eq:requirement},

\begin{equation} \label{eq:requirement}
    \begin{cases}
    d^A < L_{RR} = q|S|p + (1-q)|S|(1-p) < |S| \\
    0 < q < \frac{1}{2} \\ 
    \frac{1}{2} < p < 1
    \end{cases}
\end{equation}

\noindent where $q$ is the fraction of the number of \textbf{1}s in the origin $\boldsymbol{m}$, and $L_{RR}$ is the number of \textbf{1}s in the obscured vector $RR(\boldsymbol{m})$. These requirements can guarantee that Alice can't establish the relationship between training samples and residues, so Bob's private labels are protected. More detailed security analysis is provided in Sec.\ref{subsubsec:sec-anal}.

The parameter setting of the protocol and the detailed training time comparison among $\mathcal{M}_{add}$, $\mathcal{M}_{mult}$, $\mathcal{M}_{hybrid}$ ,and the baseline protocol in \cite{lr-no-third-party} is given in Sec.\ref{exp:hybrid}.

\subsubsection{Security Analysis} \label{subsubsec:sec-anal}
The training protocol and the corresponding transmitted messages are shown in Table \ref{tab:hybrid}.
% In the training protocol as shown in Table \ref{tab:hybrid}, the intermediate messages from Bob to Alice are listed in Table \ref{tab:inter-msgs}. 
Since in our adversary model we assume Alice to be an honest-but-curious attacker,  she can only infer the private data of Bob via these transmitted messages. Below we prove that all these messages are safe to transmit.

% \renewcommand{\arraystretch}{1.5}
% \begin{table}[h!] 
%     \centering
%     \caption{Intermediate messages \\transmitted from Bob to Alice}
%     \label{tab:inter-msgs}
%     \begin{tabular}{c |c}
%          \toprule
%          \textbf{Steps} & \textbf{Intermediate message} \\
%          \midrule
%          Step 1 & $RR(\boldsymbol{m})$ \\
%          Step 2 & $\langle \boldsymbol{r}_1 \cup \boldsymbol{r}_2 \rangle$ \\
%          Step 3 & $\boldsymbol{g}^A + \xi$ \\
%          \bottomrule
%     \end{tabular}
% \end{table}

\textbf{Step 1.} The private indicator vector $\boldsymbol{m}$ is obscured via the random response mechanism, which is $\epsilon$-LDP, then the privacy of $\boldsymbol{m}$ is guaranteed by privacy budget $\epsilon$. So with probability $p$ closer to $\frac{1}{2}$ we can provide a stronger privacy-preserving mechanism on $\boldsymbol{m}$. To this end, Alice can not know which samples are involved in the current mini-batch, so she can not establish the linear equations like Eq.\ref{eq:linear-equations}, which means that she can not directly steal private labels via equation solving.

\textbf{Step 2.} The residue set $\boldsymbol{r}_1 \cup \boldsymbol{r}_2$ is encrypted via paillier public key, since Alice has no private key , she can not infer residue values directly via ciphertext decryption.

\textbf{Step 3.} Bob sends the decrypted masked gradient to Alice and she can obtain the true gradient after removing the mask. Due to the constraints in Eq.\ref{eq:requirement} which is that the number of \textbf{1}s in the obscured indicator vector is larger than the number of Alice's features, so Alice can only construct a linear system with unknowns more than equations, which means that she can not get the residue value by solving linear equations.

Furthermore, because parameter $k$ which is the size of set $\boldsymbol{r}_1$ is unknown to Alice, she can not know either the actual mini-batch size or which samples the mini-batch consists of. The only way Alice can construct the ground truth linear system successfully is by first enumerating all possibilities of mini-batch size and then enumerating all combinations of samples to form a mini-batch with that size. Because  on average the time complexity of solving a system of linear equations with $n$ unknowns is $\mathcal{O}(n^2)$, the time complexity of the problem Alice tries to solve (finding the ground truth linear equations) is given by

\begin{equation}
    \begin{aligned}
    \sum_{k=1}^{L_{RR}} k^2 \binom{L_{RR}}{k} &=  L_{RR} (L_{RR} + 1) 2^{L_{RR} - 2} \\
    &= \mathcal{O}(L_{RR}^2 2 ^{L_{RR}}).
    \end{aligned}
\end{equation}

\noindent The complexity of this problem is even higher than many conventional problems with exponential complexity, so we assume that if Alice's computation resources are limited, and $L_{RR}$ is large enough, Alice can not find the ground truth linear equations, so Bob's private labels are being well protected.

% \end{document}

%%%%%%%%%%%%%%%%%%
% Experiments
%%%%%%%%%%%%%%%%%%
\section{Experiments}
% \documentclass[main.tex]{subfiles}
% \begin{document}
In this section, we first introduce the dataset information and our vertical federated learning system in Sec.\ref{exp:setup}. Then in Sec.\ref{exp:add}, under different privacy budgets, we evaluate the impact of additive noise on the global model performance. Next, in Sec.\ref{exp:mult}, how clipping bounds $b_1$ and $b_2$ affect the global model is validated. We also compare the performance of  $\mathcal{M}_{add}$ and $\mathcal{M}_{mult}$ under the same privacy budget. Finally, in Sec.\ref{exp:hybrid}, we give the setting of parameters in our hybrid training protocol and assess the performance of $\mathcal{M}_{hybrid}$. The training time comparison among $\mathcal{M}_{add}$, $\mathcal{M}_{mult}$, $\mathcal{M}_{hybrid}$ ,and baseline protocol in \cite{lr-no-third-party} is also presented.

\renewcommand{\arraystretch}{1.4}
\begin{table*}[ht!] 
    \centering
    \caption{Global model performance comparison between $\mathcal{M}_{add}$ and $\mathcal{M}_{mult}$}
    \label{tab:acc-auc}
    \begin{tabular}{c| c | c  | c c c c | c c c c}
         \toprule
         \textbf{Dataset} & \textbf{Metrics} & \textbf{Baseline} & \multicolumn{4}{c|}{$\mathcal{M}_{add}$}  &  \multicolumn{4}{c}{$\mathcal{M}_{mult}$}\\
         \midrule
         
          \textemdash &  \textemdash & \textemdash & $\epsilon=0.01$ & $\epsilon=0.1$ & $\epsilon=1$ & $\epsilon=10$ & $\epsilon=0.01$ & $\epsilon=0.1$ & $\epsilon=1$ & $\epsilon=10$ \\
          \hline
          \multirow{2}{1.5cm}{breast-cancer} & Acc &  97.37 & 83.33 & 87.72 &  92.10 &  94.74 &  87.72 &  89.47 & 92.98 &  95.61 \\
          &Auc & 99.87 & 87.52 &  95.84 &  98.92 &  99.43 & 95.90 &  96.79 & 97.78 &  99.21 \\
          \hline
          \multirow{2}{1.5cm}{sklearn-digits} & Acc & 91.11 &  63.33 &  77.50 &  89.17 &  90.83 &  77.50 &  79.17 &  85.00 &   89.72 \\
          & Auc &  96.66 &  63.02 &  86.99 &  95.84 &  96.72 &  86.36 &  85.90 &  89.71 &  96.14 \\
          \hline
          \multirow{2}{1.5cm}{census-income} & Acc & 82.69 &  63.43 &  71.35 &  80.89 &  82.06 &  63.77 &  63.32 &  69.77 &  79.45 \\
          & Auc &  88.79 &  66.72 &  75.85 &   85.95 &  88.76 &  75.57 &  76.88 &  83.45 &  87.74 \\
          \hline
          \multirow{2}{1.5cm}{give-me-some-credit} & Acc &  85.56 &  71.24 & 77.44 &  82.77 &  85.54 &  63.15 &  69.51 &   71.91 &  88.14 \\
          & Auc & 81.15 &  70.50 &  76.97 &   79.24 &  81.08 & 65.21 &  71.16 &  76.61 &  81.44\\
         \bottomrule
    \end{tabular}
    
\end{table*}

\subsection{Experimental Setup} \label{exp:setup}

\subsubsection{Datasets} We evaluate the effectiveness of the three proposed mechanisms via four different datasets which are \textit{breast-cancer}\footnote{\href{https://bit.ly/3sKQe8m}{https://scikit-learn.org/stable/modules/generated/breast-cancer}}, \textit{sklearn-digits}\footnote{\href{https://bit.ly/3oXT2xE}{https://scikit-learn.org/stable/modules/generated/load-digits}}, \textit{census-income}\footnote{\href{http://archive.ics.uci.edu/ml/datasets/Census+Income}{http://archive.ics.uci.edu/ml/datasets/Census+Income}}, \textit{give-me-some-credit}\footnote{\href{https://www.kaggle.com/c/GiveMeSomeCredit}{https://www.kaggle.com/c/GiveMeSomeCredit}}, respectively. We summarize the characteristics of these four datasets in Table \ref{tab:dataset-stat}.

Specifically, (1) \textit{breast-cancer}: It contains 569 samples and each sample has 30 numerical features. It is used to predict whether a person has breast cancer or not. (2)\textit{ sklearn-digits}: The original sklearn-digits dataset consists of 1,797 gray-scale images with a size of 8$\times$8 and is for multi-label classification task. We first flatten image into a vector of size 64, then we group images with an odd label into one class and all others into another class, so that we can train a binary classifier. (3)\textit{ census-income}: This dataset contains 48,842 samples and each sample is composed of 14 categorical and numerical features. We do feature engineering on it and finally get a dataset with 81 features, and it is used to predict whether a person could make \$50K a year. (4) \textit{ give-me-some-credit}: There are 150,000 samples and each one with a feature size of 10. It is used to predict whether a financial institution will loan money to a person.

\renewcommand{\arraystretch}{1.5}
\begin{table}[h!]
    \centering
    \caption{Datasets Characteristics}
    \label{tab:dataset-stat}
    \begin{tabular}{l|l|l|p{1.2cm}}
         \toprule
         \textbf{Dataset} & \textbf{\# Samples} & \textbf{\# Features} & \textbf{Task} \\
         \midrule
         breast-cancer & 569 & 30 & \multirow{4}{2cm}{Binary Classification} \\
         \cline{1-3}
         sklearn-digits & 1,797 & 64  \\
         \cline{1-3}
         census-income & 48,842 & 81  \\
         \cline{1-3}
         give-me-some-credit & 150,000 & 10 \\
         \bottomrule
    \end{tabular}
\end{table}

\subsubsection{Vertical Federated Learning System} We deploy the two parties Alice and Bob on two cloud machines with memory size 64GB. The two parties use remote procedure call framework, $\mathtt{gRPC}$\footnote{\href{https://github.com/grpc/grpc}{https://github.com/grpc/grpc}}, to communicate with each other across the Internet, and use open-source package $\mathtt{python}$-$\mathtt{paillier}$\cite{PythonPaillier} to implement Paillier cryptosystem. In our system, we allocate all features to Alice and only labels to Bob, so that we can evaluate the impact of extra noise on global model performance under such an extreme case, that's because all gradients will be affected by the extra noise. 

\begin{figure*}[ht!] 
	\begin{center}
		\begin{minipage}[t]{\linewidth}
			\centering
			\subfigure[breast-cancer]{\label{b1-breast-cancer}
				\includegraphics[width=0.23\linewidth, ]{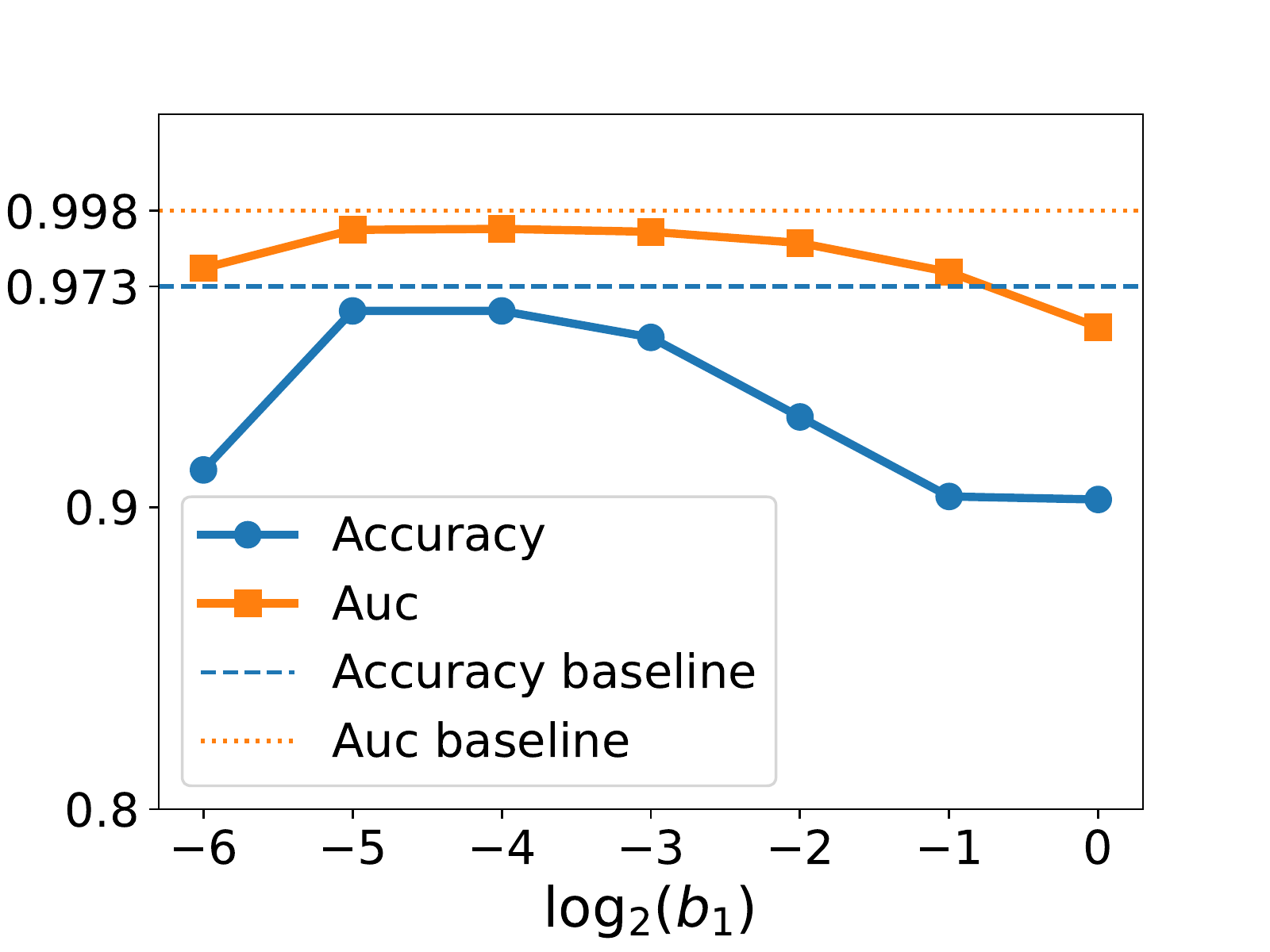}}
			\subfigure[sklearn-digits]{\label{b1-sklearn-digits}
				\includegraphics[width=0.23\linewidth ]{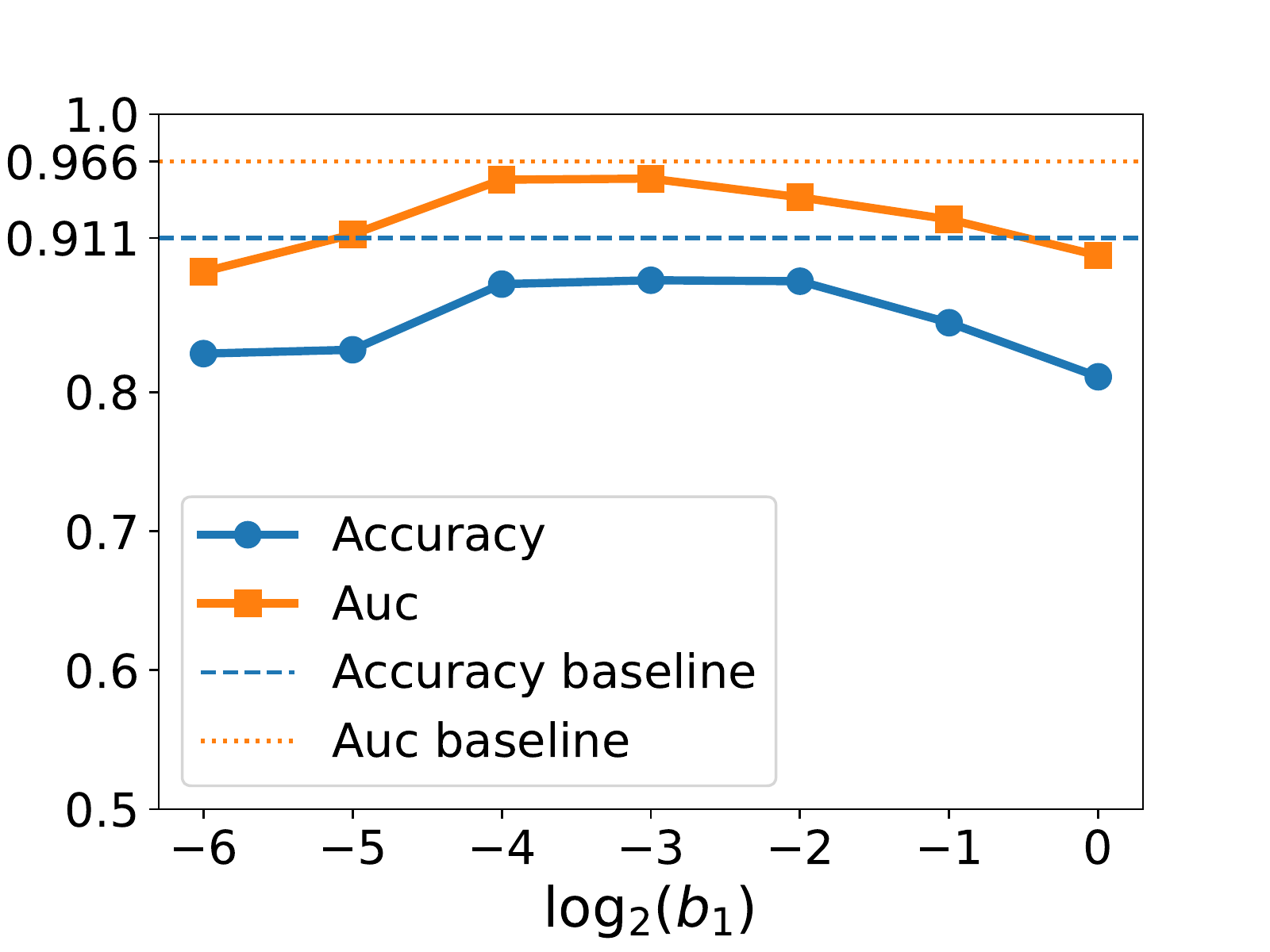}}
			\subfigure[census-income]{\label{b1-census-income}
				\includegraphics[width=0.23\linewidth ]{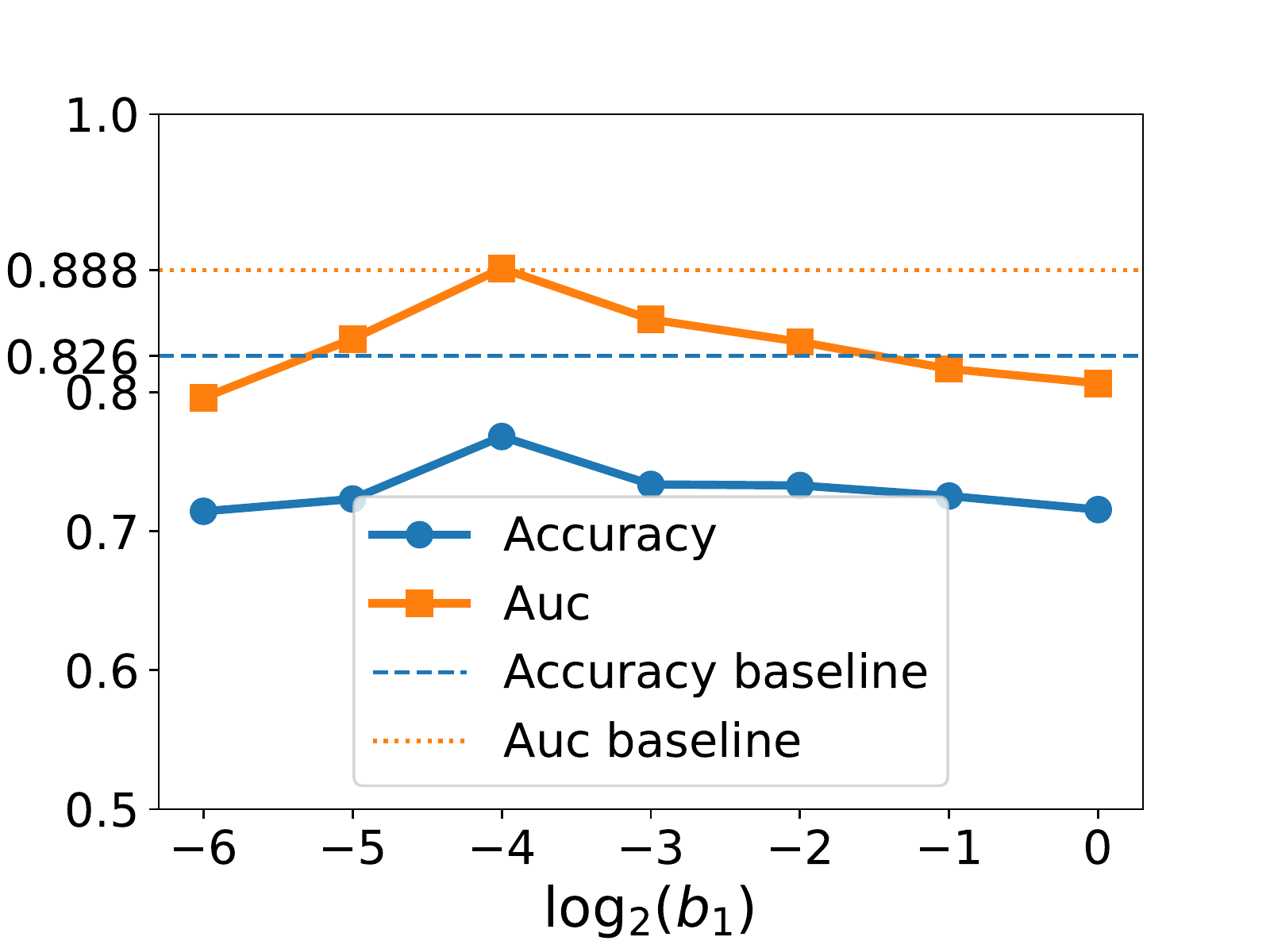}}
			\subfigure[give-me-some-credit]{\label{b1-credit}
				\includegraphics[width=0.23\linewidth ]{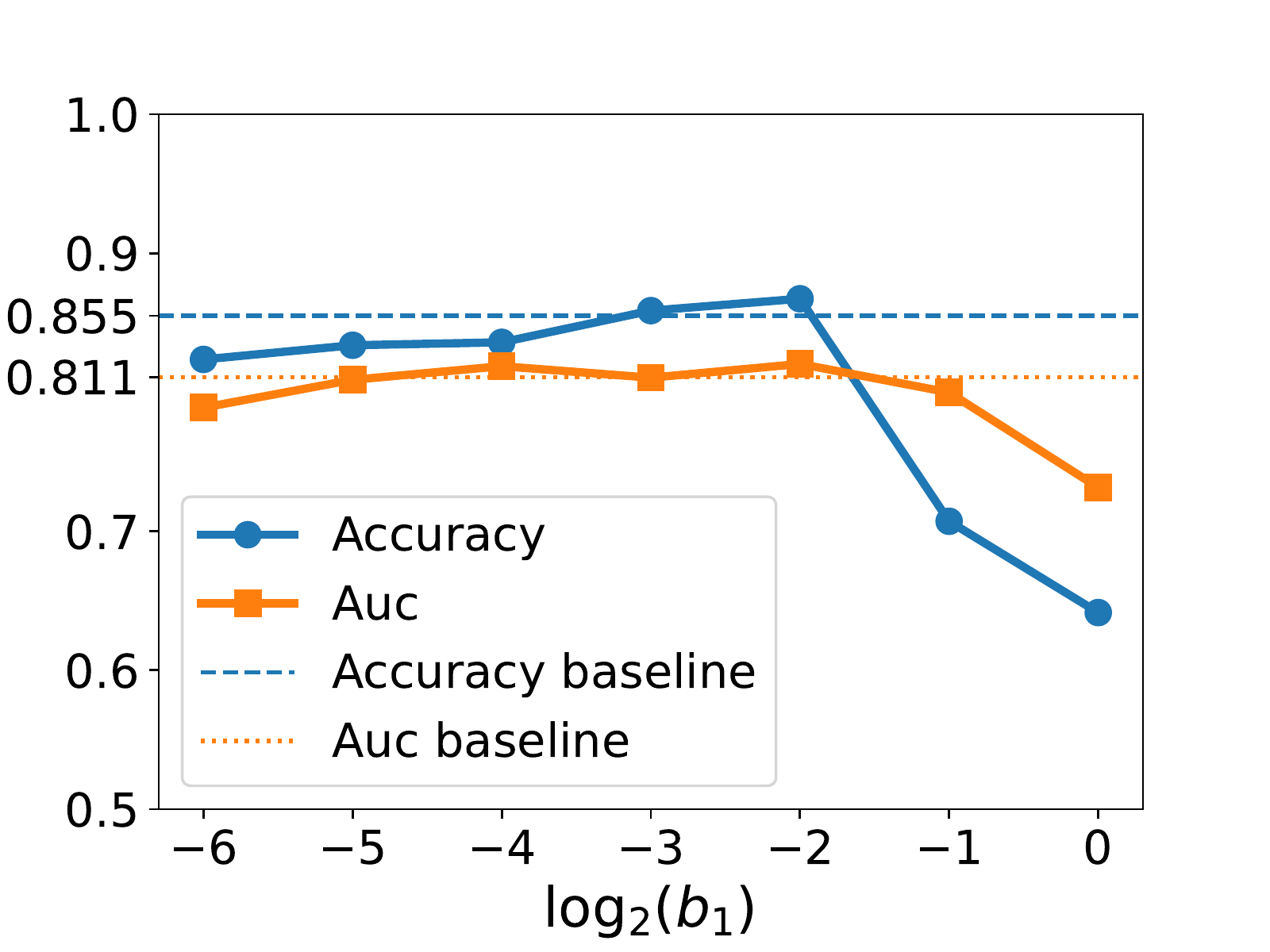}}
			\caption{Global model performance \textit{w.r.t.} $b_1$, where we fix $b_2=10$ and $\epsilon=10$.}
			\label{fig:b1-vary}
		\end{minipage}
	\end{center}
	\vspace{-0.3in}
\end{figure*}

\begin{figure*}[ht!] 
	\begin{center}
		\begin{minipage}[t]{\linewidth}
			\centering
			\subfigure[breast-cancer]{\label{b2-breast-cancer}
				\includegraphics[width=0.23\linewidth ]{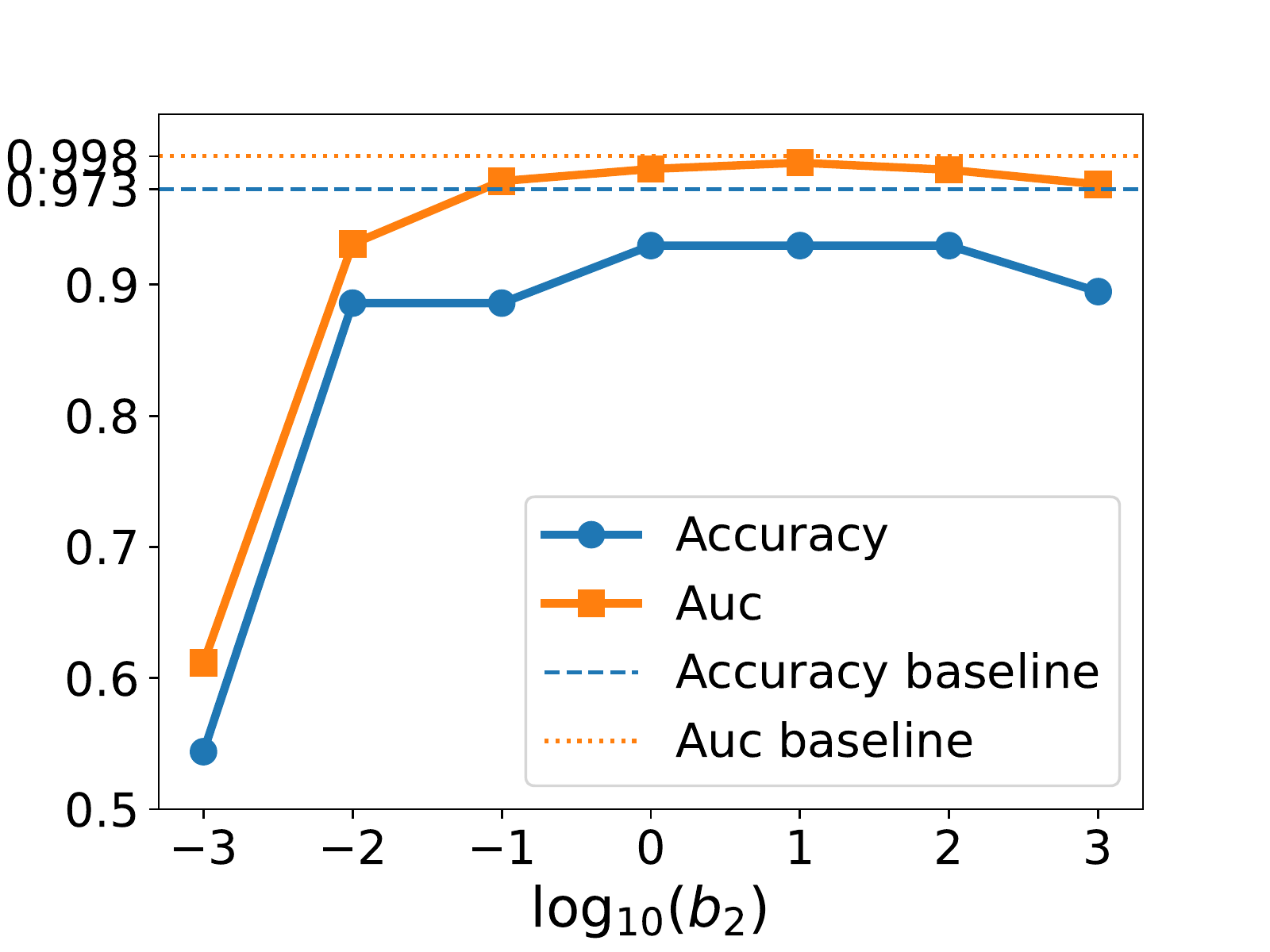}}
			\subfigure[sklearn-digits]{\label{b2-sklearn-digits}
				\includegraphics[width=0.23\linewidth ]{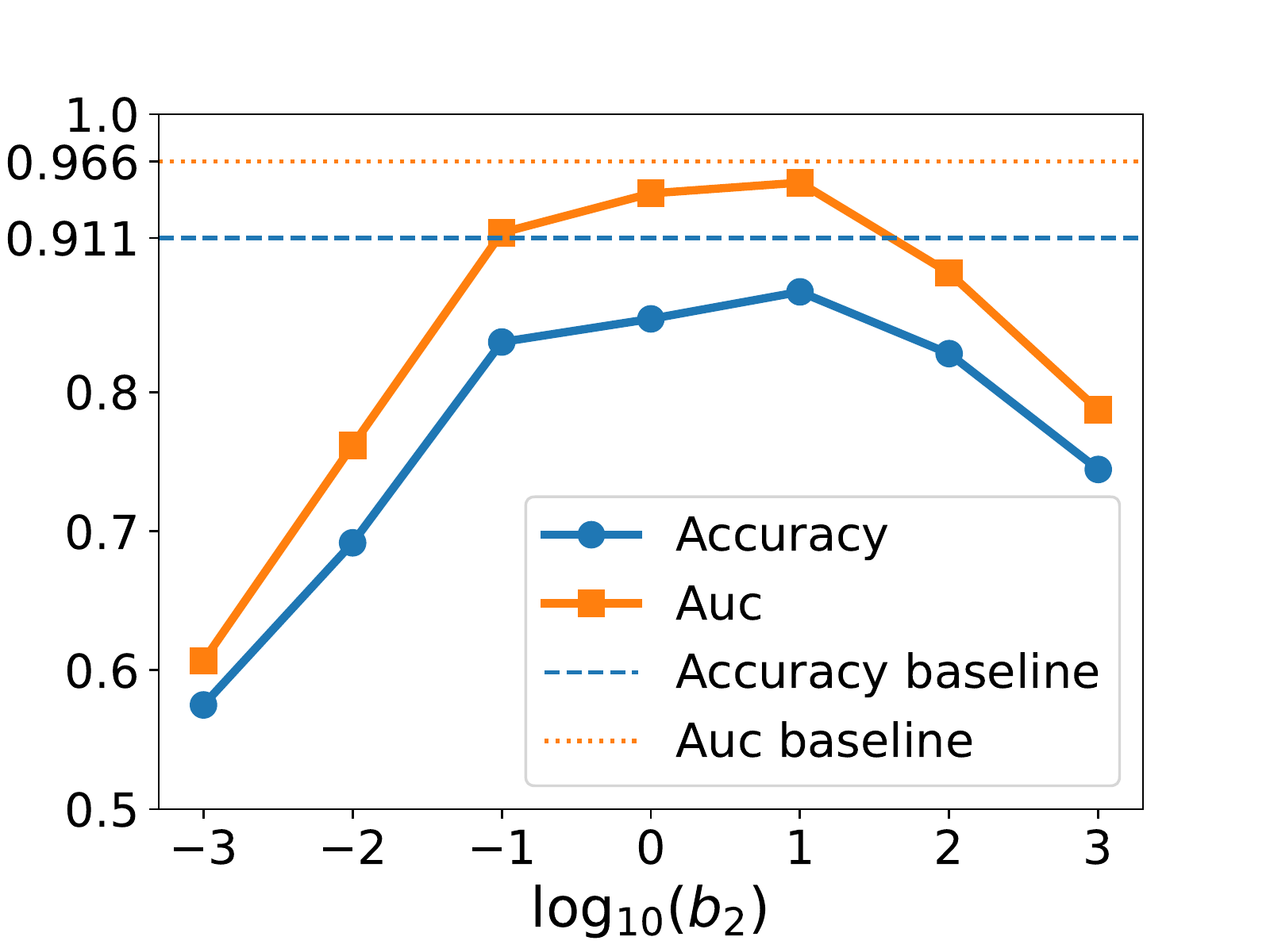}}
			\subfigure[census-income]{\label{b2-census-income}
				\includegraphics[width=0.23\linewidth]{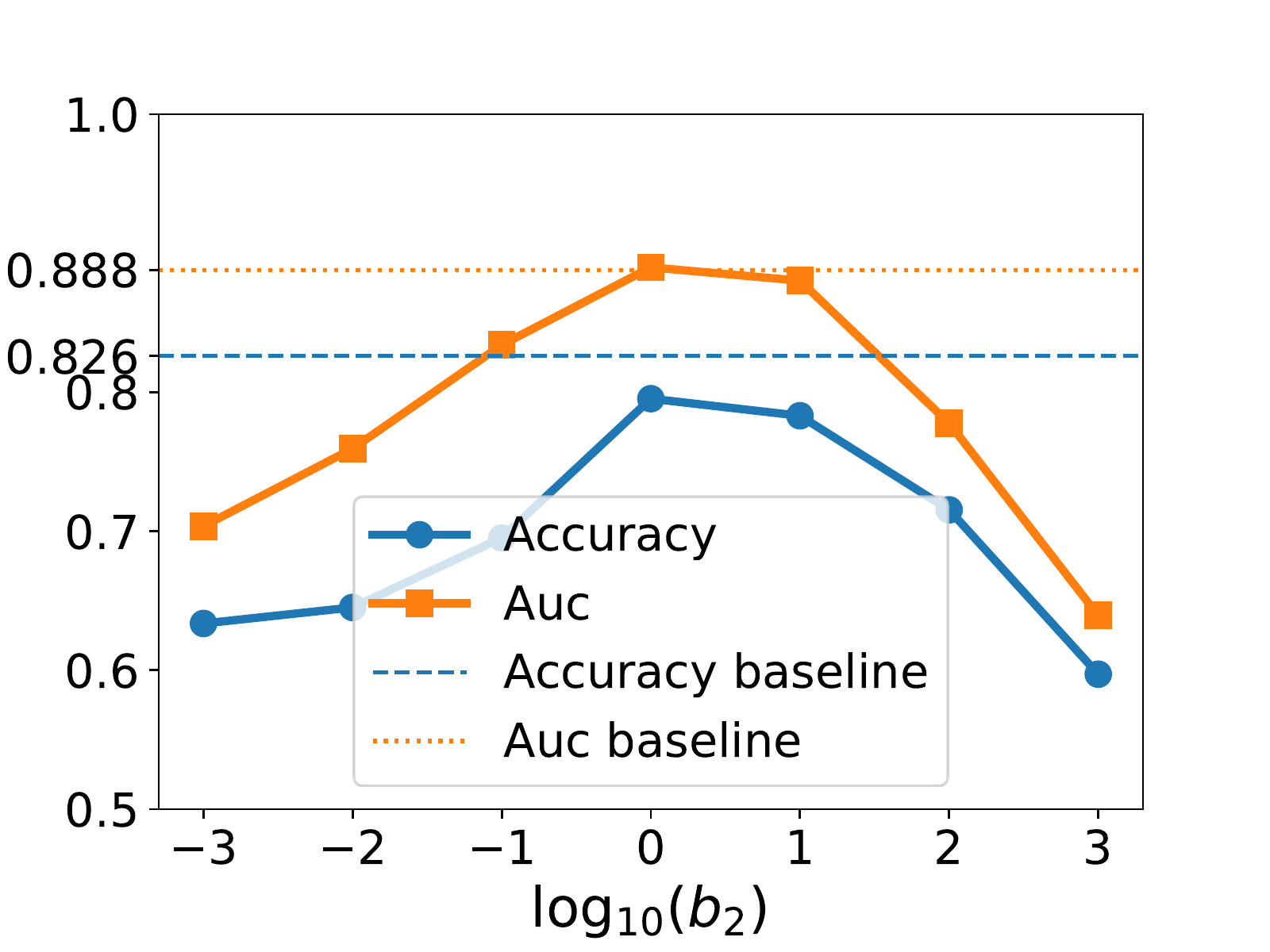}}
			\subfigure[give-me-some-credit]{\label{b2-credit}
				\includegraphics[width=0.23\linewidth]{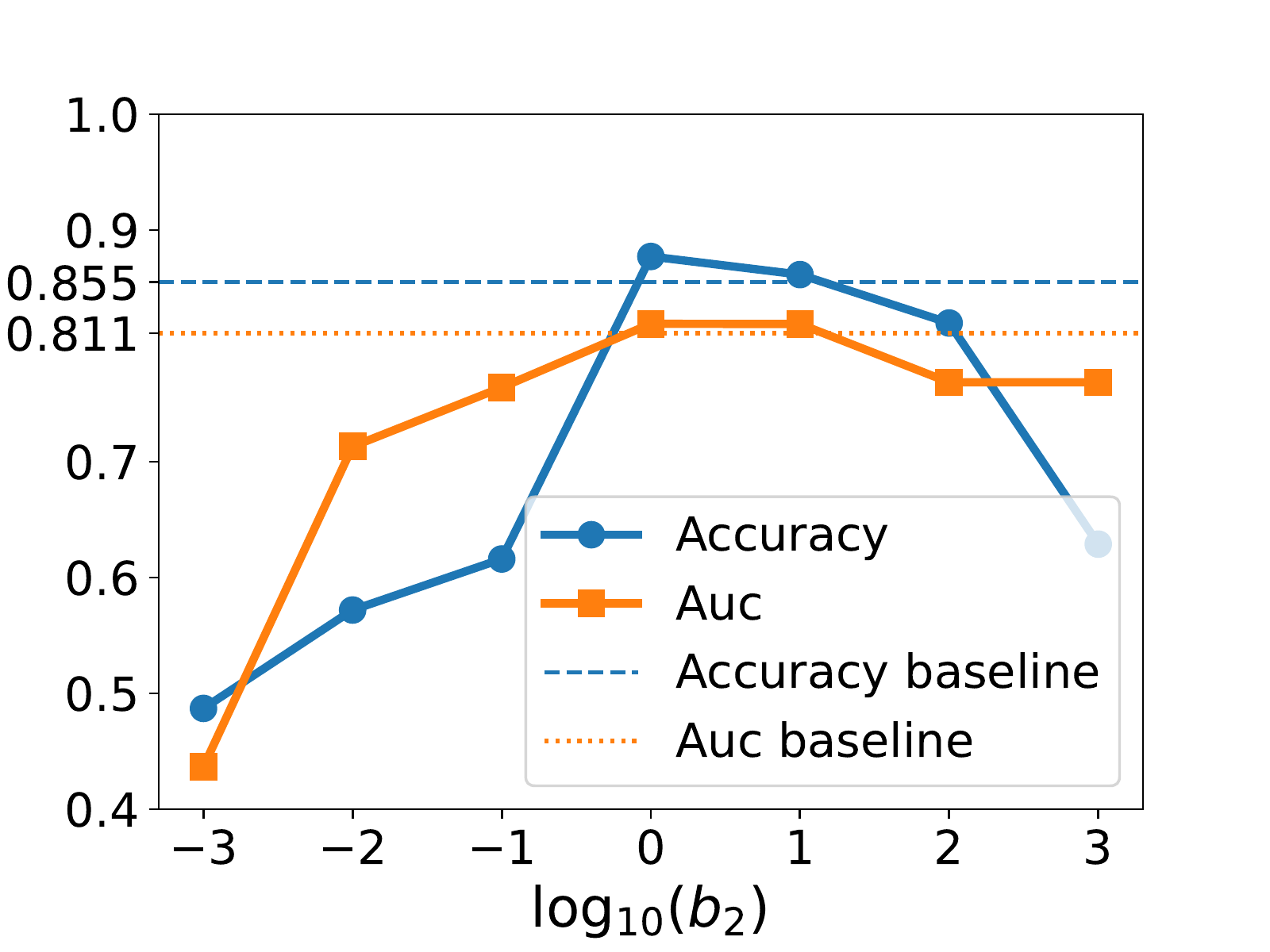}}
			\caption{Global model performance \textit{w.r.t.} $b_2$, where we fix $b_1=0.1$ and $\epsilon=10$.}
			\label{fig:b2-vary}
			\vspace{0.1in}	
		\end{minipage}
	\end{center}
	\vspace{-0.3in}
\end{figure*}

\begin{figure*}[ht!] 
	\begin{center}
		\begin{minipage}[t]{\linewidth}
			\centering
			\subfigure[breast-cancer]{\label{eps-breast-cancer}
				\includegraphics[width=0.23\linewidth ]{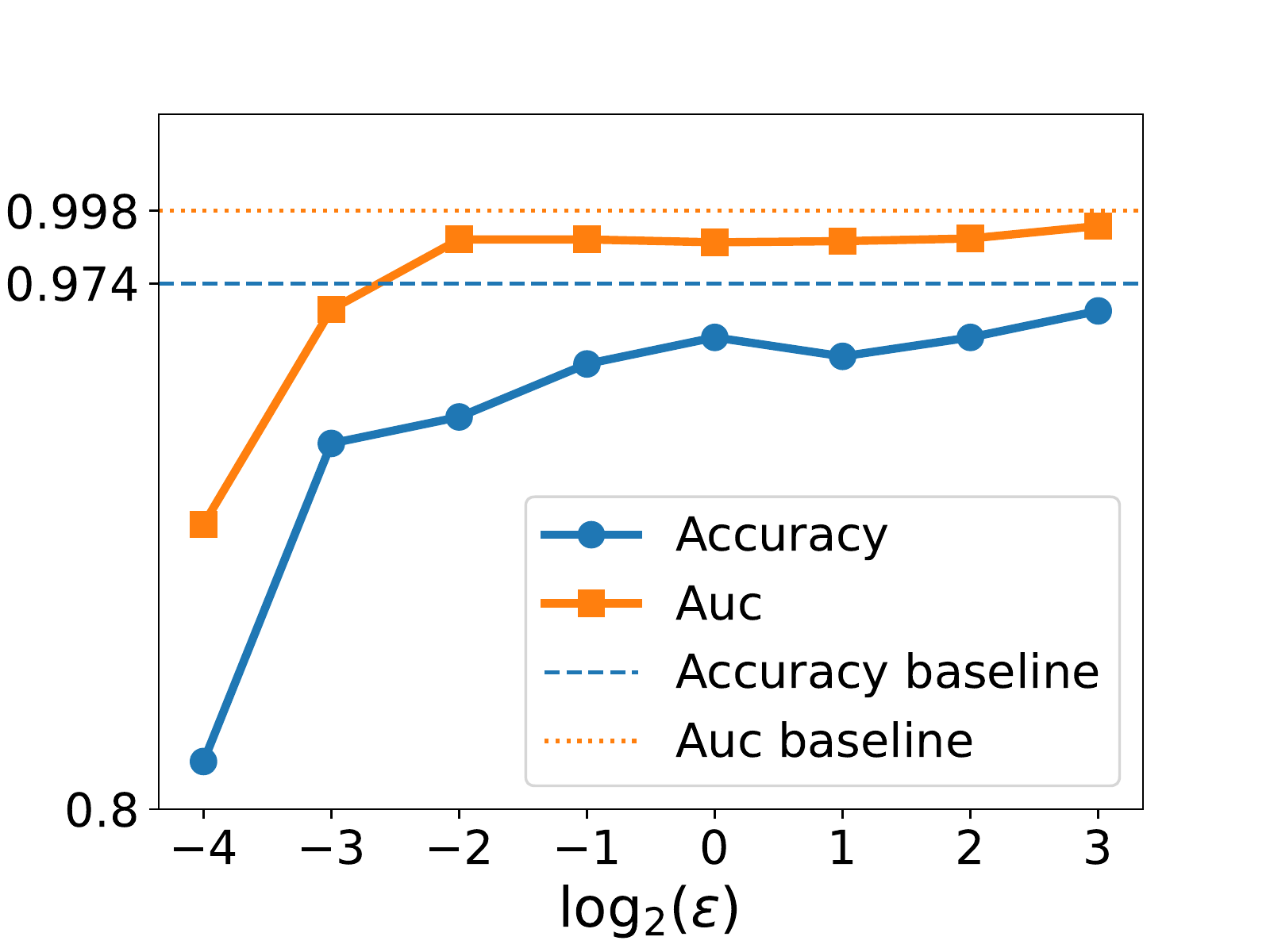}}
			\subfigure[sklearn-digits]{\label{eps-sklearn-digits}
				\includegraphics[width=0.23\linewidth ]{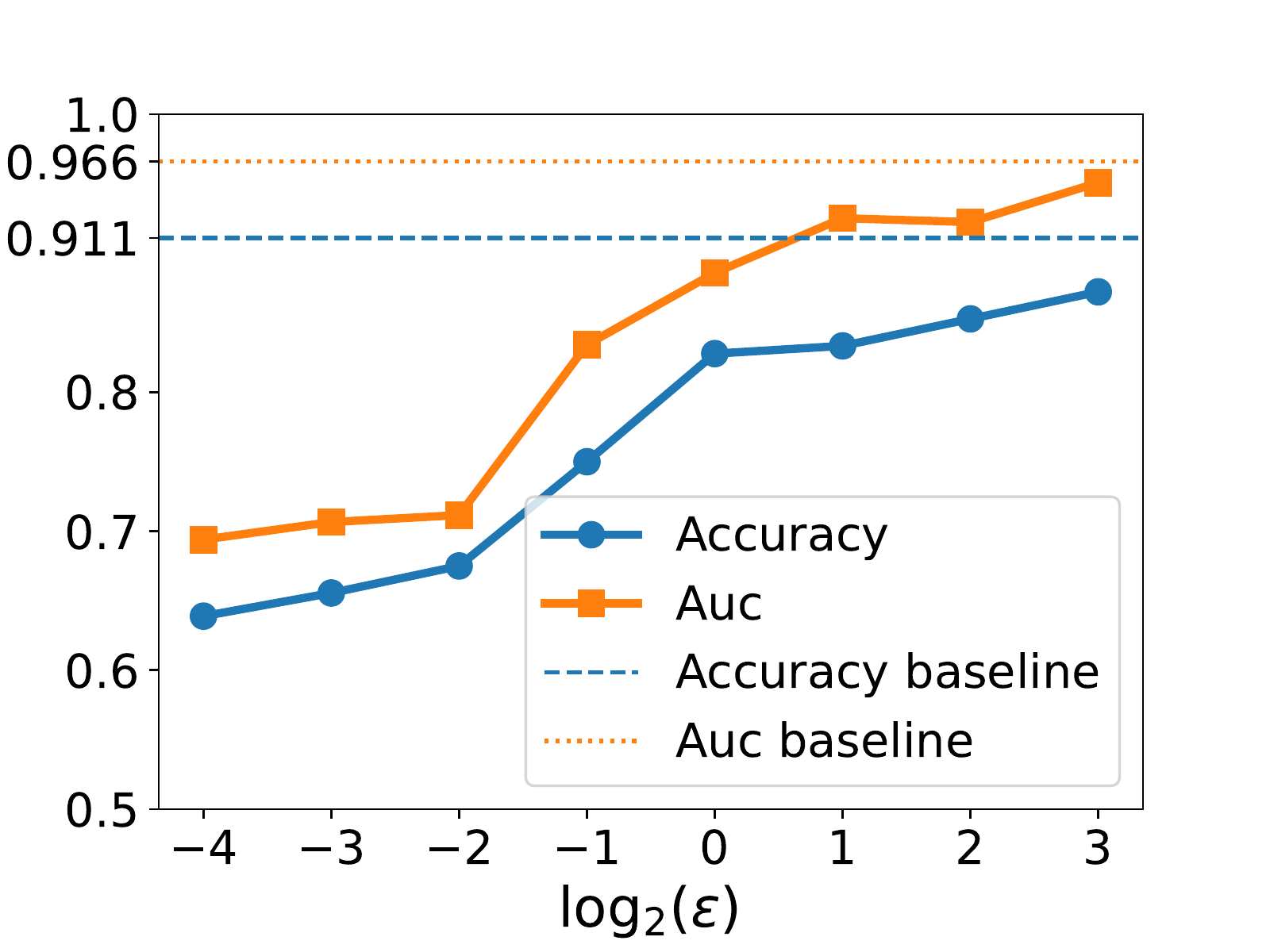}}
			\subfigure[census-income]{\label{eps-census-income}
				\includegraphics[width=0.23\linewidth]{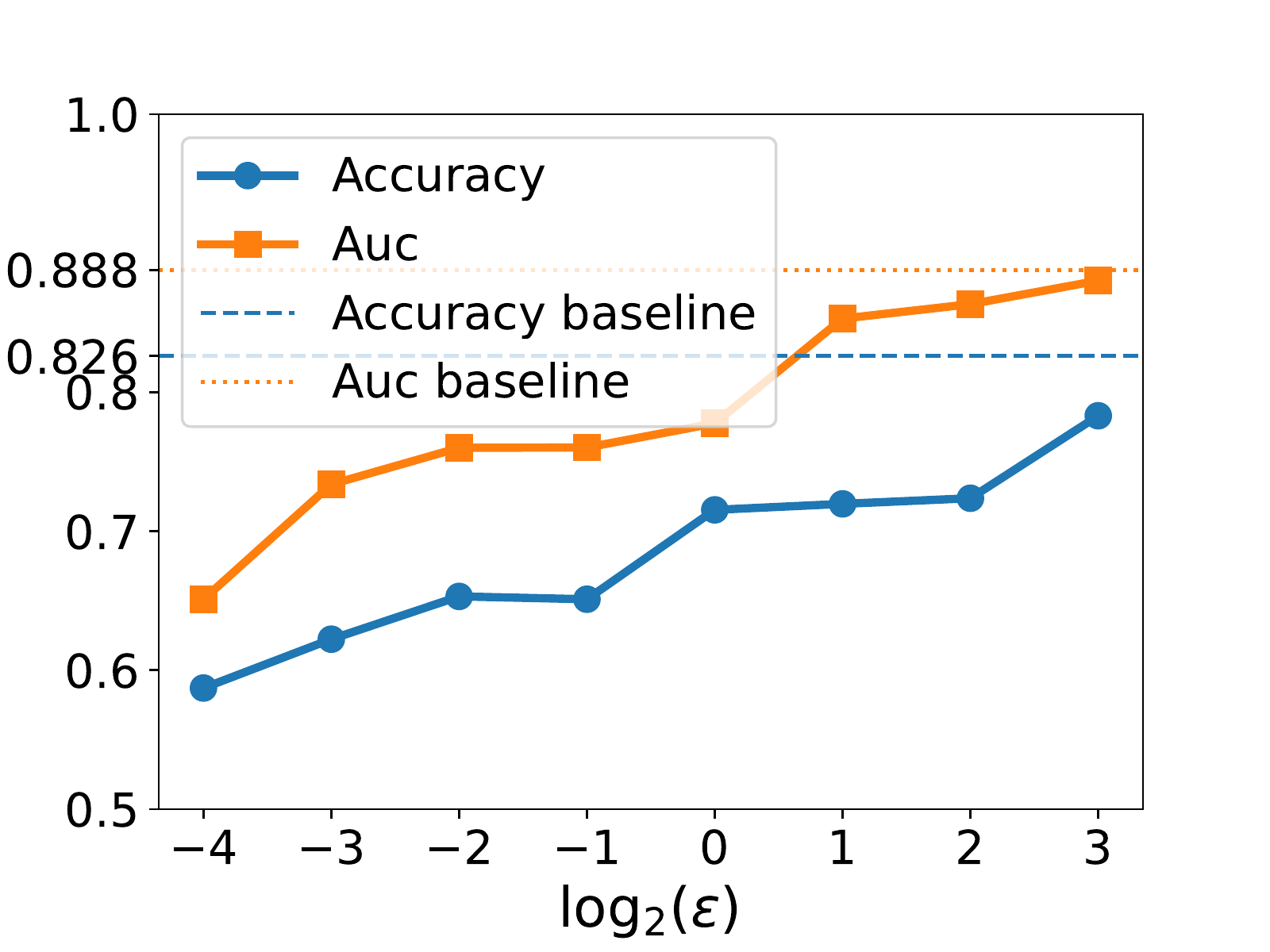}}
			\subfigure[give-me-some-credit]{\label{eps-credit}
				\includegraphics[width=0.23\linewidth]{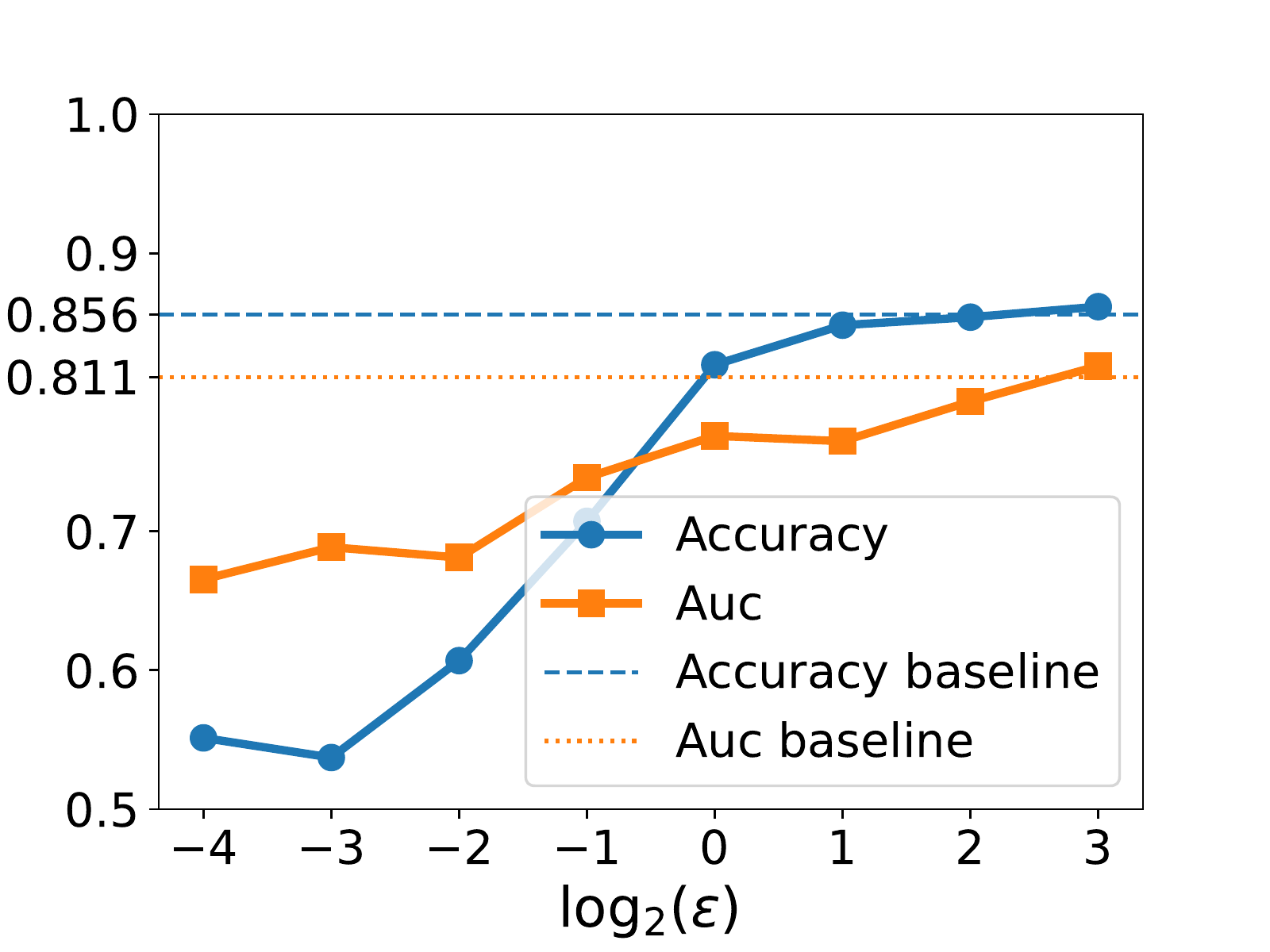}}
			\caption{Global model performance \textit{w.r.t.} $\epsilon$, where we fix $b_1=0.1$ and $b_2=10$.}
			\label{fig:eps-vary}
			\vspace{0.1in}	
		\end{minipage}
	\end{center}
	\vspace{-0.3in}
\end{figure*}

\subsection{Evaluation of $\mathcal{M}_{add}$} \label{exp:add}

Table \ref{tab:acc-auc} shows the accuracy and AUC score of the global model under different privacy budget $\epsilon$, and the validation results of a model which is trained in a centralized manner are also presented as a baseline. For all four datasets, we can see that the model performance decreases as we decrease $\epsilon$, this is in line with what we talk about in Sec.\ref{mechanism-add}, since in mechanism $\mathcal{M}_{add}$, we use a Laplace distribution $\text{Lap}(\frac{2}{\epsilon})$ with variance $2(\frac{2}{\epsilon})^2$ to generate noise, a smaller $\epsilon$ means that we ensure a stronger privacy guarantee for residue, but we get a noise distribution with larger variance, which affects the integrity of residue set and further the performance of the global model.

If the required privacy strength is not that high, \textit{e.g.}, $\epsilon=10$, from Table \ref{tab:acc-auc} we can see that on dataset breast-cancer, sklearn-digits, census-income, and give-me-some-credit, when compared with the baseline, the accuracy of the global model only decrease 0.026, 0.003, 0.006, and 0.0001, which is really small, and the AUC score of global model merely decrease 0.004, -0.0006, 0.0003, 0.0003, which means that in sklearn-digits dataset, the additive noise could even improve the AUC score. We presume that since the variance of the added noise is small, the additive noise has little negative impact on the integrity of the residue set, and in contrast, it improves the robustness of the global model.

\subsection{Evaluation of $\mathcal{M}_{mult}$} \label{exp:mult}

As mentioned in Sec.\ref{mechanism-mult}, we multiply the residue by Laplace noise with a variance $2(\frac{2b_2}{b_1\epsilon})^2$, which is determined by $b_1$, $b_2$ , and $\epsilon$ together. So in this section, we investigate the impact of these parameters on the global model performance separately.

\subsubsection{Impact of clipping bound $b_1$} \label{b1-impact}
We fix both $b_2$ and $\epsilon$ to 10 and vary $b_1$ from $2^{-6}$ to $2^0$ to observe its impact. Fig.\ref{fig:b1-vary} shows the accuracy and AUC score of the global model under different $b_1$. We observe that the  trend in all four datasets is the same, and that is both the model accuracy and AUC score first increase and then decrease with $b_1$. We think the reason behind this is:
\begin{itemize}
    \item With a small $b_1$ although we have less possibility to trigger clipping method in Eq.\ref{eq:clip1}, but the $L_1$ sensitivity calculated on residue set is big, so the variance of the multiplicative noise is large and we will trigger clipping method in Eq.\ref{eq:clip2} with high possibility. As a result, many residue items will be mapped to $b_2$ by $\mathcal{M}_{mult}$, which are not informative enough for model training.
    \item With a large $b_1$ although the variance of the noise is not large,  many residue items will be clipped to $b_1$, which causes the loss of integrity of the original  residue set, and this is also not good for model training.
\end{itemize}

So we can draw a conclusion that there is a trade-off between the integrity of the original residue set and the variance of noise. Both a small $b_1$ and a large $b_1$ are not good enough to train a good global model. There exists an optimal $b_1$ between the small one and the large one, and this explains why the curve in Fig.\ref{fig:b1-vary} increases first and then decreases with $b_1$.

\begin{table*}[ht!] 
    \centering
    \caption{Training time (seconds) comparison among $\mathcal{M}_{add}$, $\mathcal{M}_{mult}$, $\mathcal{M}_{hybrid}$ and  \cite{lr-no-third-party}}
    \label{tab:comp-hybrid}
    \begin{tabular}{c | c c c | c c|  c c  c}
         \toprule
         \textbf{Dataset} & \textbf{\# Alice's Features} & \textbf{Batch Size} $ $& $L_{RR}$ &  $\mathcal{M}_{add}$(s) & $\mathcal{M}_{mult}$(s) & \cite{lr-no-third-party}(s) & $\mathcal{M}_{hybrid}$(s) & \textbf{ratio} \\
         \midrule
         breast-cancer & 30 & 16  & 40 & 3.5 & 3.5 & 554.2 & 957.4 & 1.73\\
         sklearn-digits & 64 & 32 & 70 &  5.6 & 5.7 & 2,415.8 & 4,142.5 & 1.71 \\
         census-income & 81 & 32 & 90 & 100.5 & 101.8 & 74,049.6 & 131,299.8  &  1.77\\
         give-me-some-credit & 10 & 8 & 15 & 1,374.3 & 1,402.1 &  88,321.9 & 113,761.6 & 1.29\\
         
         \bottomrule
    \end{tabular}
\end{table*}

\subsubsection{Impact of clipping bound $b_2$}
Similarly, we fix $b_1 = 0.1$ and $\epsilon = 10$ and vary $b_2$ from 0.001 to 1,000 to observe its effect on global model performance. Fig.\ref{fig:b2-vary} shows that model accuracy and AUC score also increase first and then decrease with $b_2$. As in Sec.\ref{b1-impact}, we also give the reason as below:
\begin{itemize}
    \item With a small $b_2$ although the variance of noise is small, there is a high probability that the output of $\mathcal{M}_{mult}$ will trigger the clipping method in Eq.\ref{eq:clip2}, which causes the information loss of residue set. As a result, the accuracy and AUC score are not high.
    \item A large $b_2$ results in a large noise variance, and the original residue set with each item $-1< r< 1$ is mapped to a new set composed of very large items which cause the loss of model accuracy and AUC score, though these large items trigger Eq.\ref{eq:clip2} with a low probability.
\end{itemize}

So we also draw a conclusion that there is a trade-off between the noise variance and the integrity of the new residue set after $\mathcal{M}_{mult}$, and this trade-off is controlled by parameter $b_2$. Both a small $b_2$ and a large $b_2$ cause the loss of model performance severely. 

Notice that on dataset census-income and give-me-some-credit, when $b_2 = 1$ the AUC score is even higher than baseline, we presume this is also because the extra noise improves the model robustness as Sec.\ref{exp:add}.

\subsubsection{Impact of $\epsilon$}
Fig.\ref{fig:eps-vary} shows the model accuracy and AUC score \textit{w.r.t.} privacy budget from $2^{-4}$ to $2^3$ on four datasets, where we set $b_1=0.1$ and $b_2 = 10$ respectively. Similarly, we see a trade-off between model utility and privacy strength, where with a smaller $\epsilon$ we provide a stronger privacy guarantee for residue, but the performance of the global model is worse.

\subsubsection{Comparison with $\mathcal{M}_{add}$}
In this section, we compare the performance of the two mechanisms $\mathcal{M}_{add}$ and $\mathcal{M}_{mult}$ under the same privacy budget. The noise variance in $\mathcal{M}_{add}$ is only determined by $\epsilon$, but in $\mathcal{M}_{mult}$, it is determined by $\epsilon$, $b_1$, and $b_2$ together. So after searching for many combinations of $b_1$ and $b_2$, only the best model accuracy and AUC score of $\mathcal{M}_{mult}$ we found are reported in Table \ref{tab:acc-auc}, and we get two main observations from it:
\begin{itemize}
    \item When $\epsilon$ is small, the performance of $\mathcal{M}_{mult}$ is better than $\mathcal{M}_{add}$. For example, when $\epsilon= 0.01$ both model accuracy and AUC score in $\mathcal{M}_{mult}$ are higher than that in $\mathcal{M}_{add}$ on datasets breast-cancer, sklearn-digits, and census-income. Also, when $\epsilon=0.1$, the accuracy in $\mathcal{M}_{mult}$ on breast-cancer and sklearn-digits dataset is higher than $\mathcal{M}_{add}$. We think this is because the noise variance in $\mathcal{M}_{add}$ is really large, the original value in the residue set is overwhelmed by such noise. But in $\mathcal{M}_{mult}$ due to the clipping method in Eq.\ref{eq:clip2}, the item in the obscured residue set is no larger than that in $\mathcal{M}_{add}$, so the model performance is better.
    \item When $\epsilon$ is large, $\mathcal{M}_{add}$ performs better than $\mathcal{M}_{mult}$, which is because the noise variance in $\mathcal{M}_{add}$ is small and the impact of the extra noise on model performance is negligible. But in $\mathcal{M}_{mult}$, due to the constraints of two clipping bounds $b_1$ and $b_2$, there is a large loss of integrity of residue set so that the model performance is worse.
\end{itemize}

Therefore, users can choose to use $\mathcal{M}_{add}$ or $\mathcal{M}_{mult}$ according to their privacy budget. We empirically enumerate different $b_1$ and $b_2$ to find the best model performance, how to search for the optimal clipping bounds is out of scope in this paper, we leave it for future research. 

\subsection{Evaluation of $\mathcal{M}_{hybrid}$} \label{exp:hybrid}
As mentioned in Sec.\ref{mechanism-hybrid}, the training protocol of $\mathcal{M}_{hybrid}$ is not only secure but also lossless, because we set the residue of samples that are not in the ground-truth mini-batch to zero, these samples have no impact on the gradient. At each iteration, the gradient is kept intact so the global model performance is lossless when compared with the centralized one. Since these samples are also used during model training, which can cause the increase of protocol running time. So in this section, we investigate how much extra training time does $\mathcal{M}_{hybrid}$ bring when compared with conventional protocol in \cite{lr-no-third-party}.

Table \ref{tab:comp-hybrid} shows the setting of key parameters and the corresponding training time on four datasets. In all cases, the number of Alice's features is larger than the batch size, if we train vertical LR model via protocol in \cite{lr-no-third-party}, Bob's private labels are leaked. So we set $L_{RR}$ larger than the number of Alice's features in $\mathcal{M}_{hybrid}$ such that Bob's label is protected. The last column shows the ratio of the computation time of $\mathcal{M}_{hybrid}$ to that of the protocol in \cite{lr-no-third-party}, we observe that although we set $L_{RR}$ larger than the number of Alice's features which is equivalent to enlarge mini-batch size at each iteration, the ratio does not exceed 1.8 times on all four datasets, and this overhead is acceptable. 

We also observe that the training time of $\mathcal{M}_{add}$ and $\mathcal{M}_{mult}$ is much less than $\mathcal{M}_{hybrid}$ and the protocol in \cite{lr-no-third-party}, the reason is that there are no encryption, decryption, addition on ciphertext and  multiplication on ciphertext operations in these two mechanisms, but the disadvantages are that the extra noise decreases the model performance. So depending on users' requirements, if they need a performance-lossless protocol, then $\mathcal{M}_{hybrid}$ is a good choice, or else if they prefer an efficient protocol, then it's suitable to choose either $\mathcal{M}_{add}$ or $\mathcal{M}_{mult}$.

% \end{document}

\label{sec:exp}
% \subfile{experiments}

%%%%%%%%%%%%%%%%%%
% Conclusion
%%%%%%%%%%%%%%%%%%
\section{Conclusion}
% \subfile{conclusion}
% \documentclass[main.tex]{subfiles}
% \begin{document}

In this paper, we first present one label inference attack method to reveal the vulnerability of the widely used training protocol of vertical logistic regression.
It shows that the attacker can utilize the residue variables to infer the privately owned labels. 
Then, we propose three residue protection mechanisms, \emph{e.g.}, additive noise mechanism, multiplicative noise mechanism, and the hybrid mechanism which leverages LDP and HE techniques, to prevent the attack and improve the robustness of the vertical logistic regression model. 
Finally, we conduct comprehensive experiments to evaluate the effectiveness and efficiency of these three mechanisms. 
The results show that both the additive noise mechanism and the multiplicative noise mechanism can achieve efficient label protection with only a minor decrease of model performance in the case that the privacy budget is relatively high, and the hybrid mechanism can achieve label protection without any model accuracy degradation. 
In addition, the computation overhead of the hybrid mechanism is no more than 1.8 times that of the widely used vertical LR training protocol in \cite{lr-no-third-party}, which is usually acceptable in practice . 
\section*{Acknowledgement}
% \subfile{ack}
% \documentclass[main.tex]{subfiles}
% \begin{document}

Lan Zhang is the corresponding author. This research is supported by the National Key R\&D Program of China 2021YFB2900103, China National Natural Science Foundation with No. 61932016, No. 62132018, No. 61822209, Key Research Program of Frontier Sciences, CAS. No. QYZDY-SSW-JSC002. This work was partially supported by“the Fundamental Research Funds for the Central Universities”.

% \end{document}

%%%%%%%%%%%%%%%%%%
% References
%%%%%%%%%%%%%%%%%%
\bibliographystyle{IEEEtran}
\bibliography{reference}

\end{document}